\documentclass{article} % For LaTeX2e
\usepackage{colm2024_conference}

\usepackage{microtype}
\usepackage{hyperref}
\usepackage{url}
\usepackage{booktabs}
\usepackage{microtype}
\usepackage{cuted}   % For the strip environment
\usepackage{afterpage}
\usepackage{xcolor,colortbl}
\usepackage{mdframed}
\usepackage{amssymb}
\usepackage{listings}
\usepackage{graphicx}
\usepackage{subcaption}
\usepackage{enumitem}
\usepackage{verbatim}
\usepackage{caption}

\colmfinalcopy

\newmdenv[
  backgroundcolor=gray!20, % Set background color to light gray
  linewidth=0pt,            % No border
  innerleftmargin=10pt,     % Padding inside the frame
  innerrightmargin=10pt,
  innertopmargin=10pt,
  innerbottommargin=10pt
]{customquote}

\newcommand{\nw}[1]{}

\title{Fusion-Eval: Integrating Assistant Evaluators with LLMs}

\author{
Lei Shu\textsuperscript{1}\thanks{Correspondence to leishu@google.com .} \qquad
Nevan Wichers\textsuperscript{1} \qquad
Liangchen Luo\textsuperscript{1} \qquad
Yun Zhu\textsuperscript{1} \qquad
Yinxiao Liu\textsuperscript{1} \AND
Jindong Chen\textsuperscript{1} \qquad
Lei Meng\textsuperscript{2} \\ \\
Google Deepmind\textsuperscript{1}, Google\textsuperscript{2}
}

\begin{document}
\maketitle
\begin{abstract}
Evaluating natural language systems poses significant challenges, particularly in the realms of natural language understanding and high-level reasoning. 
In this paper, we introduce ``Fusion-Eval'', an innovative approach that leverages Large Language Models (LLMs) to integrate insights from various assistant evaluators. The LLM is given the example to evaluate along with scores from the assistant evaluators. Each of these evaluators specializes in assessing distinct aspects of responses. % This unique strategy enables Fusion-Eval to function effectively across a diverse range of tasks and criteria, enhancing the effectiveness of existing evaluation methods.
Fusion-Eval achieves a 0.962 system-level Kendall-Tau correlation with humans on SummEval and a 0.744 turn-level Spearman correlation on TopicalChat, which is significantly higher than baseline methods. 
These results highlight Fusion-Eval's significant potential in the realm of natural language system evaluation.
\end{abstract}

\section{Introduction}
\label{sec:introduction}
Evaluating the performance of natural language generation models has significant challenges~\citep{ouyang2022training}, particularly in terms of evaluation benchmarks and evaluation paradigms~\citep{wang2023aligning}. 
This study focuses on the latter one.
Typically, the evaluation paradigms fall into three categories: human-based, automatic-metrics-based and model-based evaluations.
Among these, human evaluations are regarded as the most reliable, yet they come with high costs and issues of scalability.

Automatic metrics such as BLEU~\citep{papineni2002bleu} and ROUGE~\citep{lin2004rouge} are prevalent in evaluations, relying on comparisons with a `gold' standard reference. However, the creation of these gold references is a labor-intensive process. % Moreover, in tasks involving content generation, the variety of potential correct responses can mean that comparisons to a single or limited number of references may not fully capture the quality of the generated content. 
Furthermore, studies such as \citet{fabbri2021summeval} have demonstrated that these automatic metrics often do not correlate well with human judgment.

Model-based evaluations aim to enhance the correlation with human judgment using neural networks fine-tuned on specific datasets. Neural evaluators like BLEURT~\citep{sellam2020bleurt} and its variant SMART~\citep{amplayo2022smart} show improved alignment with human assessments in various generative tasks. These models offer flexibility in evaluation methods. They can either compare the response to the source (reference-free), or to the gold standard (reference-dependent).

Recent advancements have seen the use of Large Language Models (LLMs) as reference-free evaluators in Natural Language Generation (NLG) tasks. Notably, studies by \citet{fu2023gptscore, wang2023chatgpt} have leveraged LLMs to rate candidate outputs based on their generation probability alone, eliminating the need for reference text comparisons. Additionally, \citet{liu2023gpteval} introduced a method called G-Eval, where LLMs, guided by human-crafted evaluation criteria, score responses. Meta-evaluations indicate that these LLM-based evaluators reach a level of human correlation on par with medium-sized neural evaluators~\citep{zhong2022towards}.
In light of these developments in evaluation paradigms, the following question arises:
\begin{customquote}
``Can Large Language Models (LLMs) integrate existing evaluators to achieve higher correlation with human judgments?''
\end{customquote}
% devise an evaluation plan and

In response to this question, we introduce \textit{Fusion-Eval}, an innovative evaluation framework that integrates a variety of existing evaluators—termed \textit{assistant evaluators}—to enhance correlation with human judgment. Fusion-Eval prompts an LLM with an example to evaluate and scores given by assistant evaluators. In our work, we consider reference free evaluation. Fusion-Eval can evaluate any natural language task where assistant evaluators are available. However, its effectiveness hinges on the quality of the assistant evaluators, making it more suitable for well-established text generation tasks.

\section{Method}
\label{sec:method}

Fusion-Eval is an evaluation framework leveraging a Large Language Model (LLM) to fuse assistant evaluators, to improve scoring quality.
The framework's goal is to evaluate a Natural Language Generation (NLG) system along one or more criteria in a manner highly correlated with human judgment. The test examples are what Fusion-Eval will evaluate. For example in the SummEval dataset, a test example is a news article and a summary. In this cause, Fusion-Eval will evaluate the quality of the summary given the news article. % For examples, see Section \ref{sec:experiment_setting} and Figure~\ref{fig:fusioneval_schema}.
Each assistant evaluator receives a test example and returns a score. The Fusion-Eval framework then takes evaluation task descriptions, test examples, and assistant evaluator scores as inputs.
We propose two Fusion-Eval solutions:

\paragraph{(1) Fusion-Eval without Plan (FE-NoPlan)}
In this method, the Large Language Model (LLM) is prompted directly with the task's evaluation criteria, details about assistant evaluators, and a request for evaluation scores. This prompt also includes placeholders for the assistant evaluator scores and the test example, as well as instructions on the format the LLM should use to generate the evaluation scores. This straightforward approach requires the LLM to interpret the evaluation criteria and information on assistant evaluators without a predefined plan. Table~\ref{tab:prompt} presents a simplified prompt template for Fusion-Eval without Plan (FE-NoPlan).
\begin{table}[h]
\centering
% \small
\begin{tabular}{|p{13cm}|}
\hline
\\
\textsl{You are an evaluation agent. I will give you one summary written for a news article. Please evaluate the quality of the summary.} \\\\
\textsl{Detailed descriptions of these metrics are as follows:}\\\\
\textsl{Coherence(1-5, Any Floating Value):the collective quality of all sentences. \textless{}...\textgreater{}}\\\\
\textsl{Three assistant evaluators are provided.}\\\\
\textsl{1. Natural Language Inference (NLI) provides the probability of the entailed relationship between source text (as premise). Its range is between 0-1, close to 1 indicates that the hypothesis is entailed by the premise.\textless{}...\textgreater{}}\\\\
\textsl{Use these evaluators as supplementary tools for your judgement and rate the responses across the five metrics \textless{}...\textgreater{}}\\
\\
% \nw{We should also include the part where we give the model the actual summary to evaluate. One of the reviwers was confused about this.}
Input Template: \textless{}...\textgreater{}\\
\\
Output Template: \\
Coherence Score:  [Your  evaluation ]  Explanation :  [Your  explanation  on  evaluation ] \textless{}...\textgreater{}\\
\\
Input Example: \\
Source: \\
\{source\} \\
\\
Answer: \\
\{summary\} \\
\\
NLI Score (Source as Premise and Answer as Hypothesis):\\
\{nli\}\\
\\
BLEURT Score (Source as Premise and Answer as Hypothesis):\\
\{bleurt\}\\
\\
SUM\_BLEURT Score (Source as Premise and Answer as Hypothesis):\\
\{sumbleurt\}\\
\\
Evaluation (please follow Output Template and provide the evaluation result):
\\
\hline
\end{tabular}
\caption{Trimmed Prompt for Fusion-Eval without Plan for the SummEval dataset.}
\label{tab:prompt}
\end{table}
% Please share your understanding of the evaluation task and plan for using assistant evaluators, including criteria planning and steps. <...>

\paragraph{(2) Fusion-Eval with Plan (FE)}
\begin{figure}[h] % The [h] here is a placement specifier that suggests placing the figure "here"
  \centering % This centers the figure
  \includegraphics[width=0.96\textwidth]{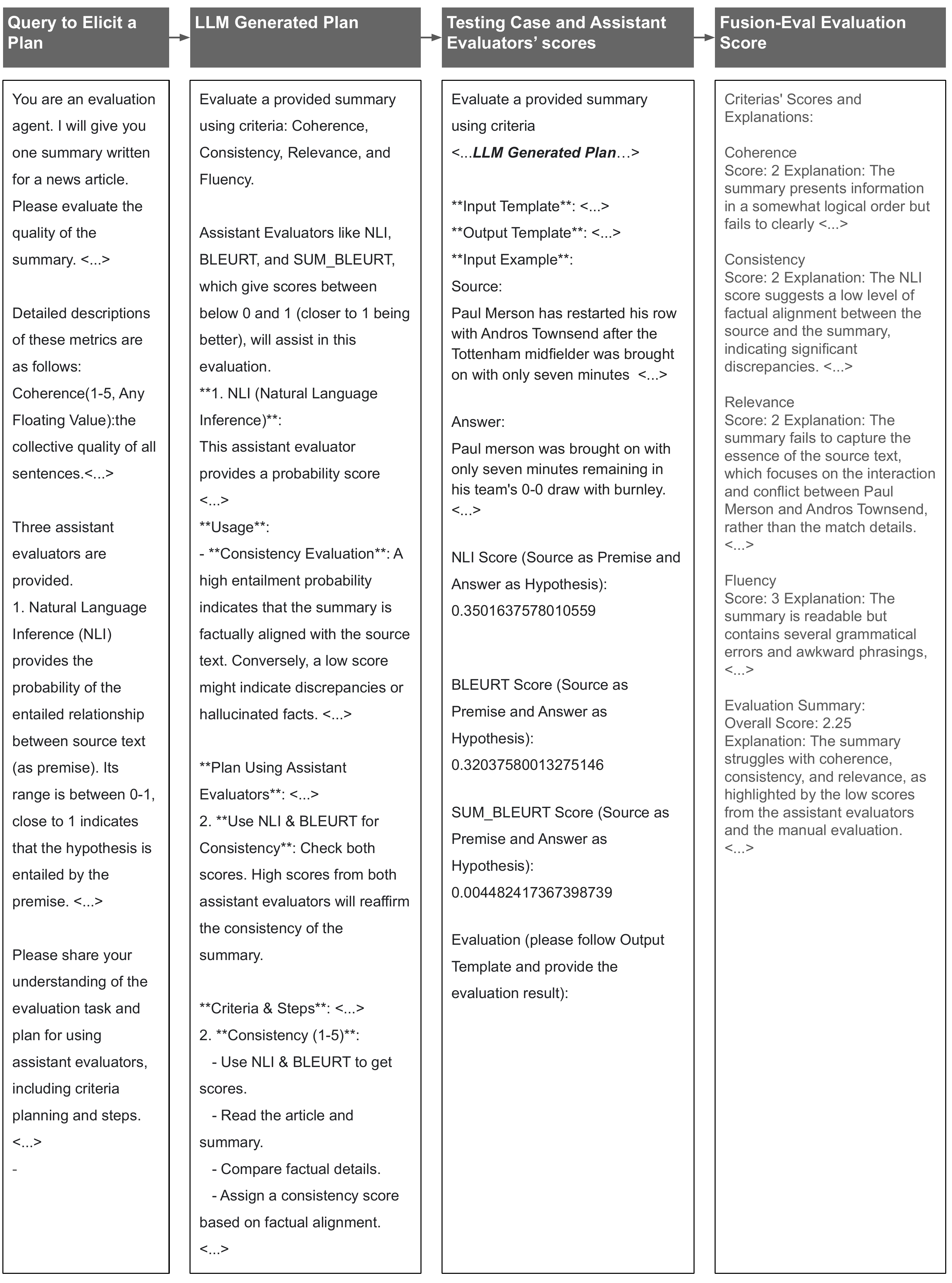} % Adjust the path and the width as needed
  \caption{Workflow of Fusion-Eval with Plan: Starting from the left, a query initiates the generation of a plan by the LLM. Once the plan is obtained, it is concatenated with the template. The template placeholders are filled in for each test example along with its specific assistant evaluators' scores. This complete prompt is then used to obtain the Fusion-Eval evaluation score from the LLM. \nw{Make the text at the top of this figure bigger.}} % Caption of the figure
  \label{fig:fusioneval_schema} % Label for referencing the figure in text
\end{figure}

This approach introduces a plan that specifies which assistant evaluators to use for evaluating each specific criteria, accompanied by detailed steps for the LLM to follow when evaluating the test example.
It is designed for complex evaluation tasks that benefit from  guidance. The plan also adds transparency as one can see which evaluators are used for what purpose. % We insert the plan into the prompt detailed in the previous section. 
There are trade-offs between using a human-generated or an LLM-generated plan and our framework accommodates both options. While human-authored plans tend to be more accurate, those generated by LLMs offer greater scalability and faster adaptation to new evaluation tasks. This paper showcases the Fusion-Eval with Plan (FE), utilizing plans generated by an LLM.

When using an LLM to generate the plan, the LLM is prompted with the task's definition, criteria, and information about assistant evaluators. This is similar to the auto chain-of-thought method in G-Eval~\citep{liu2023gpteval}, but it uniquely incorporates assistant evaluators.
% In our experiments we use a plan generated by an LLM, as shown in Figure~\ref{fig:fusioneval_schema}.
The workflow of Fusion-Eval with Plan is illustrated in Figure~\ref{fig:fusioneval_schema}, encompassing an auto chain-of-thought process~\citep{liu2023gpteval}. Initially, we create a prompt (the leftmost textbox in Figure~\ref{fig:fusioneval_schema}) to solicit a plan from the LLM. The second textbox shows a trimmed LLM-generated plan (comprehensive plans with templates are available in Appendices~\ref{sec:appendix:prompt_summeval} and \ref{sec:appendix:prompt_topicalchat}). \nw{Commenting this out because it's similar to the sentence at the begining of this section.} % This plan outlines the LLM's understanding of the evaluation task and provides recommendations for assistant evaluators' utilization alongside a detailed evaluation criteria. 

Once we obtain the plan, we insert it into the prompt described in the FE-NoPlan section. % Upon obtaining the LLM-generated plan, we integrate it with human-created templates, incorporating the test case and their corresponding assistant evaluators' scores.
This forms the complete prompt for deriving the Fusion-Eval final score, depicted in the third textbox in Figure\ref{fig:fusioneval_schema}.

%Table~\ref{tab:llm_suggested_assistant_evaluator} shows the assistant evaluators which the LLM plans to use for each criteria. \nw{I think we should remove this part. Because when we include it, we get questions about why each assistant evaluator should be used for each criteria. TODO(me)       }

%The condensed strategic evaluation plan from the LLM is below. 
To adapt Fusion-Eval to a different evaluation task, one needs to update the criteria and assistant evaluator descriptions and regenerate the plan. Additionally, collecting new assistant evaluator scores for the task is necessary.
Full Fusion-Eval templates are available in Appendix~\ref{sec:appendix:prompt_summeval} for SummEval and \ref{sec:appendix:prompt_topicalchat} for TopicalChat.

Our framework is compatible with many possible plans, as long as they describe a valid way to incorporate the assistant evaluators. Finding the optimal plan is outside the scope of our work. 
% \nw{What do you think of this part I added?}
% the goal of our work is to show that incorporating assistant evaluator scores 

\paragraph{Prompt Execution} In both solutions, the prepared evaluation prompt template is used with each test example. This template is filled with the inputs, responses, and assistant evaluator scores for each test example. The executing LLM then processes this filled prompt, yielding Fusion-Eval's final evaluation scores as shown in the rightmost textbox in Figure~\ref{fig:fusioneval_schema}. % In our experiments, we use the scores output by the LLM directly, however FuisonEval could also work with the Scoring Function proposed in G-Eval.
We found that the LLM generated evaluation scores in the correct format, so we did not need to do anything else to control the outputs.

The executing LLM processes the complete prompt and generates a numerical score for each evaluation dimension. The LLMs are configured to produce 8 predictions with temperatures of 0.5 for PaLM2 and 0.1 for GPT-4. The final Fusion-Eval scores are the average of 8 predictions. We do this because we can't obtain log probabilities from the GPT API.

% \begin{lstlisting}
% Evaluate a provided summary using criteria: Coherence, Consistency, Relevance, and Fluency.
% Assistant Evaluators like NLI, BLEURT, and SUM_BLEURT, which give scores between below 0 and 1 (closer to 1 being better), will assist in this evaluation.
% **1. NLI (Natural Language Inference)**:
% This assistant evaluator provides a probability score indicating how much the summary (hypothesis) is entailed by the original news article (premise).
% **Usage**:
% - **Consistency Evaluation**: A high entailment probability indicates that the summary is factually aligned with the source text. <...>
% **Plan Using Assistant Evaluators**:
% 1. **Read the News Article and Summary**: <...>
% 2. **Use NLI & BLEURT for Consistency**: <...>
% **Criteria & Steps**: <...>
% 2. **Consistency (1-5)**:
%   - Use NLI & BLEURT to get scores.
%   - Read the article and summary.
%   - Compare factual details.
%   - Assign a consistency score based on factual alignment. <...>
% **Evaluation Summary (1-5)**:
% Consider the scores from each criterion and their importance. <...>
% \end{lstlisting}
% \vspace{-.7em}

%\vspace{-1mm}

\section{Experiment}
\label{sec:experiment}
We conduct a meta-evaluation of Fusion-Eval, utilizing the SummEval \citep{fabbri2021summeval} and TopicalChat \citep{mehri2020usr} benchmarks. We chose SummEval and TopicalChat as benchmarks for meta-evaluation because UniEval~\citep{zhong2022towards} and G-Eval~\citep{liu2023gpteval} also use only those benchmarks. This facilitates effective comparison with their results. These benchmarks are widely recognized and offer a comprehensive range of evaluation metrics. We intentionally excluded datasets that rely on single-rater annotations~\citep{stiennon2020learning, bai2022training} or are limited to a singular metric~\citep{wang2020asking}.

\subsection{Experiment Setting}
\label{sec:experiment_setting}
SummEval~\citep{fabbri2021summeval}, a benchmark for text summarization evaluation, consists of 1600 data points. Each data point includes average ratings from three experts on a scale of 1 to 5, spanning four summary quality dimensions: coherence (Coh), consistency (Con), fluency (Flu) and relevance (Rel). The ``Overall'' score is derived as an average across these four dimensions.
% \nw{Is the fact that there are 16 systems improtant to the eval? If not I think it should be removed to avoid confusion}\ls{removed}

TopicalChat~\citep{mehri2020usr}, a benchmark for evaluating knowledge-based dialogue response generation, includes 360 data points. It features human evaluations from three experts across six dimensions: coherence (Coh), engagingness (Eng), naturalness (Nat), groundedness (Gro), understandability (Und), and overall. Ratings for naturalness, coherence, and engagingness are on a scale from 1 to 3, while groundedness and understandability are scored between 0 and 1. The overall dimension is evaluated on a scale of 1 to 5. Each data point comprises a conversation history, a grounding fact, and a potential next-turn response.

To measure the correlation between results generated by Fusion-Eval and human evaluations, we use Kendall-Tau scores for system-level analysis in SummEval \citep{fabbri2021summeval}, and Spearman scores for turn-level analysis in TopicalChat \citep{mehri2020usr} to align with each benchmark's original scoring methodology. Although UniEval~\cite{zhong2022towards} and G-Eval~\citep{liu2023gpteval} present summary-level correlations in their papers, we derived system-level correlations from their disclosed predictions to remain consistent with SummEval's original evaluation method \citep{fabbri2021summeval}. This adjustment accounts for discrepancies between our reported scores and those initially published in the G-Eval study.

%OpenAI's GPT-4 (gpt-4) and GPT-4-turbo (gpt-4-1106-preview) 

In our experiments, PaLM2-Large \citep{anil2023palm} and GPT-4~\citep{openai2023gpt4} serve as the Large Language Models (LLMs) for execution, designated as FE-PaLM2 and FE-GPT-4, respectively. In the ablation study FE-PaLM2-NoPlan, we use the Fusion-Eval without Plan method as described in Section~\ref{sec:method}.

We integrate several assistant evaluators: NLI \citep{bowman2015large}, BLEURT \citep{sellam2020bleurt}, and SumBLEURT—a BLEURT variant fine-tuned for human summarization evaluation \citep{clark2023seahorse}. We also obtain the probability that PaLM will generate the response from the dataset given the context, following methods in \citet{fu2023gptscore} and \citet{wang2023chatgpt}. The probability of the response is higher if it's more likely according to PaLM2. We use this as an assistant evaluator called PaLM2 Prob. 

To the best of our knowledge, the LLMs used in Fusion-Eval were not trained on the SummEval and TopicalChat datasets.

\begin{table}[h]
\centering
% \fontsize{9pt}{9pt}\selectfont
% \setlength{\tabcolsep}{2pt}
\begin{tabular}{llllll}
\toprule
\rowcolor{green!20}
  & \multicolumn{5}{c}{Human Evaluation}\\
\rowcolor{green!20}  
 &
  Coh &
  Con &
  Flu &
  Rel &
  Overall \\\midrule
\multicolumn{6}{l}{\cellcolor[HTML]{DDDDDD}Reference-Based Metrics}          \\
ROUGE-1         & 0.35   & 0.55  & 0.527 & 0.583 & 0.503 \\
ROUGE-2         & 0.233  & 0.6   & 0.494 & 0.433 & 0.44  \\
ROUGE-L         & 0.117  & 0.117 & 0.259 & 0.35  & 0.211 \\
BLEU            & 0.217  & 0.05  & 0.326 & 0.383 & 0.244 \\
CHRF            & 0.35   & 0.617 & 0.561 & 0.55  & 0.519 \\
S1-CHRF         & 0.3    & 0.733 & 0.494 & 0.5   & 0.507 \\
S2-CHRF         & 0.3    & 0.7   & 0.46  & 0.433 & 0.473 \\
SL-CHRF         & 0.367  & 0.733 & 0.494 & 0.5   & 0.523 \\
BERTScore       & 0.333  & -0.03 & 0.142 & 0.2   & 0.161 \\
MoverScore      & 0.217  & -0.05 & 0.259 & 0.35  & 0.194 \\\midrule
\multicolumn{6}{l}{\cellcolor[HTML]{DDDDDD}Source-dependent Metrics} \\
BARTScore       & 0.35   & 0.617 & 0.494 & 0.45  & 0.478 \\
UniEval         & 0.683  & 0.75  & 0.661 & 0.667 & 0.728 \\
DE-PaLM2 & 0.733 & 0.6 & 0.745 & 0.85 & 0.879 \\
G-Eval (GPT-4)  & 0.733  & 0.583 & 0.778 & 0.883 & 0.912 \\ 
\midrule
\multicolumn{6}{l}{\cellcolor[HTML]{DDDDDD}Assistant Evaluators} \\
BLEURT          & 0.433  & \textbf{0.767} & 0.644 & 0.633 & 0.678 \\
NLI             & 0.45   & 0.717 & 0.628 & 0.65  & 0.695 \\
SumBLEURT       & 0.7    & 0.333 & 0.544 & 0.633 & 0.644 \\ \midrule
\multicolumn{6}{l}{\cellcolor[HTML]{DDDDDD}Aggregation of Assistant Evaluators (AE)}                           \\
AVG{\tiny(AE)}             & 0.65   & 0.55   & 0.661  & 0.783  & 0.828 \\
LLMSel{\tiny(AE)}     & 0.7    & 0.75   & -      & 0.767  & -     \\
CorrW{\tiny(AE)}  & 0.667  & 0.65   & 0.678  & 0.783  & 0.845 \\\midrule
\multicolumn{6}{l}{\cellcolor[HTML]{DDDDDD}Aggregation of AE and LLM Direct Evaluation}\\
AVG{\tiny (AE, DE-PaLM2)}                 & 0.717 & 0.583 & 0.728 & 0.85  & 0.895 \\
AVG{\tiny(AE, G-Eval-GPT-4)}             & 0.717 & 0.617 & 0.745 & 0.883 & 0.912 \\
LLMSel{\tiny(AE, DE-PaLM2)    }     & 0.733 & 0.717 & -     & 0.833 & -     \\
LLMSel{\tiny(AE, G-Eval-GPT-4) }    & 0.733 & 0.717 & -     & 0.85  & -     \\
CorrW{\tiny(AE, DE-PaLM2) }    & 0.717 & 0.633 & 0.745 & 0.85  & 0.895 \\
CorrW{\tiny(AE, G-Eval-GPT-4)} & 0.733 & 0.633 & 0.762 & 0.883 & 0.912 \\\midrule
\multicolumn{6}{l}{\cellcolor[HTML]{DDDDDD}Fusion-Eval} \\
FE-PaLM2-NoPlan & 0.767 & 0.617 & 0.728 & 0.867 & 0.895 \\
FE-PaLM2 & \textbf{0.783} & \textbf{0.767} & 0.778 & \textbf{0.917} & \textbf{0.962} \\
FE-GPT-4 & \textbf{0.783} & 0.762 &\textbf{0.812} & 0.9 & 0.946 \\
\bottomrule
\end{tabular}
\caption{System-level Kendall-Tau ($\tau$) correlations of different evaluators to human judgements on SummEval benchmark. The assistant evaluators, BLEURT, NLI and SumBLEURT, treat the article as a premise and the summary as a hypothesis.}
\label{tab:summeval_system_tau}
\end{table}
%\vspace{-1mm}

\begin{table}[h]
\centering
% \fontsize{9pt}{9pt}\selectfont
% \setlength{\tabcolsep}{1pt}
\begin{tabular}{lllllll}
\toprule
\rowcolor{green!20}
  & \multicolumn{6}{c}{Human Evaluation}\\
 \rowcolor{green!20}
 &
  Coh &
  Eng &
  Nat &
  Gro &
  Und &
  Overall \\
  \rowcolor{green!20}
  &(1-3) & (1-3) &  (1-3)& (0-1)& (0-1) & (1-5) \\\midrule
  \multicolumn{7}{l}{\cellcolor[HTML]{DDDDDD}Source-dependent Metrics} \\
UniEval    & 0.613 & 0.605 & 0.514 & 0.575 & 0.468 & 0.663 \\
DE-PaLM2& 0.669 & 0.688 & 0.542 & 0.602 & 0.493 & 0.66  \\
G-Eval (GPT-4)& 0.605 & 0.631 & 0.565 & 0.551 & - & -\\
\midrule
 \multicolumn{7}{l}{\cellcolor[HTML]{DDDDDD}Assistant Evaluators} \\
BLEURT     & 0.316 & 0.461 & 0.384 & 0.638 & 0.432 & 0.464 \\
PaLM2 Prob & 0.583 & 0.606 & 0.637 & 0.441 & 0.676 & 0.687 \\\midrule
 \multicolumn{7}{l}{\cellcolor[HTML]{DDDDDD}Aggregation of Assistant Evaluators (AE)}                                        \\
AVG{\tiny(AE)}                      & 0.556 & 0.637 & 0.626 & 0.579 & 0.672 & 0.697 \\
LLMSel{\tiny(AE)  }            & -     & -     & 0.637 & 0.638 & 0.676 & -     \\
CorrW{\tiny(AE)   }        & 0.575 & 0.637 & 0.638 & 0.6   & 0.682 & 0.703 \\\midrule
 \multicolumn{7}{l}{\cellcolor[HTML]{DDDDDD}Aggregation of AE and LLM Direct Evaluation}              \\
AVG{\tiny(AE, DE-PaLM2) }           & 0.655 & 0.708 & 0.631 & 0.639 & 0.679 & 0.737 \\
LLMSel{\tiny(AE, DE-PaLM2)}    & -     & -     & 0.637 & 0.66  & 0.68  & -     \\
CorrW{\tiny(AE, DE-PaLM2)} & 0.666 & 0.711 & 0.641 & 0.65  & \textbf{0.689} & 0.742 \\
\multicolumn{7}{l}{\cellcolor[HTML]{DDDDDD}Fusion-Eval} \\
FE-PaLM2-NoPlan & 0.683 & 0.722 & 0.649 & 0.643 & 0.641 & 0.735 \\
FE-PaLM2 & \textbf{0.697} & 0.728 & 0.651 & \textbf{0.709} & 0.632 & 0.764 \\
FE-GPT-4 &0.678 & \textbf{0.747} & \textbf{0.691} & 0.692 & 0.687 & \textbf{0.774} \\
\bottomrule
\end{tabular}
\caption{Turn-level Spearman ($\rho$) correlations of different evaluators to human judgements on TopicalChat benchmark. BLEURT treats the fact and conversation as the premise and the response as the hypothesis. PaLM2 Prob represents the conditional probability of the response given the fact and conversation. The G-Eval scores for Und and Overall are missing because they aren't reported in their paper.}
\label{tab:topicalchat_turn_spearman}
\end{table}

%\vspace{-1mm}

\begin{table}[h]
\centering
% \fontsize{9pt}{9pt}\selectfont
% \setlength{\tabcolsep}{2pt}

\begin{tabular}{lllllllllll}
\toprule

& \multicolumn{4}{c}{SummEval} & &\multicolumn{5}{c}{TopicalChat}\\

& Coh &Con &Flu &Rel &   &Coh &Eng &Nat &Gro & Und\\\midrule
BLEURT          &   & \checkmark &  &  \checkmark  & BLEURT     &  & &  & \checkmark &\\\midrule
NLI             &   & \checkmark &  &  &PaLM2 Prob &  &  & \checkmark &  & \checkmark   \\\midrule
SumBLEURT       & \checkmark   &  &  & \checkmark   &\multicolumn{6}{c}{}\\
\bottomrule
\end{tabular}
\caption{LLM-Suggested Assistant Evaluator Alignment for SummEval and TopicalChat Criteria. The criteria include coherence (Coh), consistency (Con), fluency (Flu), relevance (Rel), engagingness (Eng), naturalness (Nat), groundedness (Gro), and understandability (Und).}
\label{tab:llm_suggested_assistant_evaluator}

\begin{tabular}{llllll}
\toprule
\rowcolor{blue!10}
 &\multicolumn{5}{c}{FE-PaLM2}\\
 \rowcolor{blue!10}
  &
  Coh &
  Con &
  Flu &
  Rel & 
  Overall\\\midrule
BLEURT    & 0.583 & 0.867 & 0.733 & 0.65  & 0.717 \\
NLI     & 0.6   & 0.783 & 0.75  & 0.667 & 0.733 \\
SumBLEURT & 0.75  & 0.467 & 0.633 & 0.717 & 0.683\\
\bottomrule
\end{tabular}
\caption{FE-PaLM2 and Assistant Evaluators System-level Kendall-Tau ($\tau$) correlations on SummEval.}
\label{tab:fusioneval_assistant_evaluator_summeval_correlation}

\begin{tabular}{lllllll}
\toprule
\rowcolor{blue!10}
 & \multicolumn{6}{c}{FE-PaLM2}\\
\rowcolor{blue!10}
   &
  Coh &
  Eng &
  Nat &
  Gro &
  Und &
  Overall\\\midrule

BLEURT    & 0.524 & 0.558 & 0.59  & 0.662 & 0.622 & 0.67  \\
PaLM2 Prob & 0.711 & 0.784 & 0.808 & 0.588 & 0.711 & 0.792 \\
\bottomrule
\end{tabular}
\caption{FE-PaLM2 and Assistant Evaluators Turn-level Spearman ($\rho$) correlations on TopicalChat.}
\label{tab:fusioneval_assistant_evaluator_topicalchat_correlation}

\begin{tabular}{llllll}
\toprule
\rowcolor{blue!10}
 &\multicolumn{5}{c}{FE-GPT-4}\\
 \rowcolor{blue!10}
  &
  Coh &
  Con &
  Flu &
  Rel & 
  Overall\\\midrule
BLEURT    & 0.583 & 0.795 & 0.733 & 0.6   & 0.7   \\
NLI       & 0.633 & 0.745 & 0.717 & 0.617 & 0.717 \\
SumBLEURT & 0.717 & 0.41  & 0.633 & 0.667 & 0.667 \\
\bottomrule
\end{tabular}
\caption{FE-GPT-4 and Assistant Evaluators System-level Kendall-Tau ($\tau$) correlations on SummEval.}
\label{tab:fusioneval_gpt4_assistant_evaluator_summeval_correlation}

\begin{tabular}{lllllll}
\toprule
\rowcolor{blue!10}
 & \multicolumn{6}{c}{FE-GPT-4}\\
\rowcolor{blue!10}
   &
  Coh &
  Eng &
  Nat &
  Gro &
  Und &
  Overall\\\midrule
BLEURT     & 0.577 & 0.644 & 0.565 & 0.693 & 0.617 & 0.678 \\
PaLM2 Prob & 0.747 & 0.713 & 0.86  & 0.662 & 0.799 & 0.798 \\

\bottomrule
\end{tabular}
\caption{FE-GPT-4 and Assistant Evaluators Turn-level Spearman ($\rho$) correlations on TopicalChat.}
\label{tab:fusioneval_gpt4_assistant_evaluator_topicalchat_correlation}
\end{table}
\subsection{Baselines}
For a thorough comparison, we meta-evaluated Fusion-Eval against a range of baseline methods on the SummEval benchmark. These baselines include ROUGE \citep{lin2004rouge}, BLEU \citep{papineni2002bleu}, CHRF \citep{popovic2015chrf}, SMART \citep{amplayo2022smart}, BERTScore \citep{zhang2019bertscore}, MoverScore \citep{zhao2019moverscore}, BARTScore \citep{yuan2021bartscore}, UniEval \citep{zhong2022towards}, and G-Eval \citep{liu2023gpteval}.

\textbf{UniEval} \citep{zhong2022towards} serves as a unified multi-dimensional neural evaluator for various aspects of text generation, framing evaluation as QA tasks. It leverages a pretrained T5 model \citep{raffel2020t5} to encode the evaluation task, alongside source and target texts, in a question-and-answer format, ultimately computing the QA score as the evaluation metric. This flexibility allows it to adapt to diverse evaluation tasks through simple modifications to the question format.

\textbf{G-Eval} \citep{liu2023gpteval} leverages LLMs and chain-of-thought (CoT) reasoning to assess the quality of generated texts through a form-filling approach. By inputting only the evaluation task description and criteria into LLMs, it prompts them to create a CoT outlining detailed evaluation steps. These steps, combined with the original prompt, are then used to evaluate NLG outputs. Additionally, the probabilities associated with the output rating tokens are utilized to further refine the evaluation metric.
We derived scores for most baselines from the SMART paper \citep{amplayo2022smart}, while for UniEval\footnote{\url{https://github.com/maszhongming/UniEval}} and G-Eval\footnote{\url{https://github.com/nlpyang/geval}}, we computed system-level correlation scores from their open-access predictions to align with SummEval's evaluation framework \citep{fabbri2021summeval}, as their original publications only provided summary-level correlations.

% \nw{I still think we should give a short description of GEval and UniEval and how the methods are different from ours.}

For the TopicalChat benchmark, we compared Fusion-Eval's performance with G-Eval~\citep{liu2023gpteval} and UniEval~\citep{zhong2022towards}, utilizing scores from their respective publications. Notably, G-Eval did not report scores for the `Und' and `Overall' dimensions or predictions for the TopicalChat benchmark, so these scores are omitted from our comparison. %\nw{A reviewer was asking why some of the metrics we report for GEval are different from the g eval paper. Could you explain that here?}

We introduce DE-PaLM2 (Direct Evaluator PaLM2) as an ablation baseline, employing the same approach as G-Eval with a similar prompt. This baseline shows PaLM2's standalone performance on the SummEval and TopicalChat benchmarks without assistance from other evaluators. The designation DE-PaLM2, rather than G-Eval (PaLM2), is chosen because G-Eval's prompt for the TopicalChat benchmark was not disclosed, necessitating our own implementation of G-Eval's approach.
% We also introduce DE-PaLM2 (Direct Evaluator PaLM2) as an ablation baseline. DE-PaLM2 uses the same approach as G-Eval. Its prompt is similar to G-Eval. This baseline provides insights into PaLM2's standalone performance on the SummEval and TopicalChat benchmarks without the help from assistant evaluators.
%FE-PaLM2 but without including assistant evaluators scores, similar to G-Eval. 
%The prompt is the same as Fusion-Eval but without the sections related to assistant evaluators, and doesn't mention assistant evaluators in the plan. \nw{We need this part since it addresses a reviewer comment.}

We further propose a set of aggregation functions to merge scores from assistant evaluators: % with direct evaluations from LLMs (DE-PaLM2 and G-Eval):
\begin{itemize}
    \item \textbf{AVG (Average Scores)}: The average of the score from all evaluators.
    \item \textbf{LLMSel (LLM-Selected Assistant Evaluators)}: The average score but only from evaluators which the plan identifies as relevant to the category.
    \item \textbf{CorrW (Correlation-Weighted Average)}: The average of each evaluator score weighted by the evaluator's correlation with human judgment. 
    % \nw{From reviewer: "Another possibility for correlation-weighted average is fitting a linear regressor on train/val set. Is correlation-weighted average supposed to be better? If not why did the authors use this way of combining the assistant evaluators?"}
\end{itemize}
The AE rows, (like "AVG{\tiny(AE)}") only include the assistant evaluators in the aggregation. The rows with the name of a LLM evaluator (like "AVG{\tiny(AE, G-Eval-GPT-4)}") use both the assistant evaluator scores and the score from the LLM evaluator in the aggregation.

For SummEval, G-Eval and DE-PaLM scores (G-Eval Fluency from 1-3) were adjusted from 1-5 to a 0-1 scale to align with assistant evaluators' scoring range. For TopicalChat, our aggregation includes only assistant evaluators and DE-PaLM2, as G-Eval's predictions are unavailable. Also, DE-PaLM2's scores for coherence, engagingness, and naturalness were remapped from 1-3 to 0-1 to match the scoring ranges of BLEURT and PaLM2 Prob.

\subsection{Result Analysis}
Tables~\ref{tab:summeval_system_tau} and \ref{tab:topicalchat_turn_spearman} present the correlation of baselines, assistant evaluators, and Fusion-Eval with human judgment. 

\subsubsection{Fusion-Eval Performance}
Fusion-Eval outperforms all baseline models and aggregation methods in the overall dimension and nearly all other dimensions, as demonstrated in the FE-GPT-4 and FE-PaLM2 rows of both datasets.

The remainder of our analysis is dedicated to the overall correlation with human judgment. Among various aggregation methods for assistant evaluators, the method that weights by correlation with humans (CorrW) performs best. Aggregating the LLM direct evaluator score with assistant evaluator scores yields better results than using the direct evaluator alone for PaLM2, and it matches performance for GPT models. Specifically, AVG(AE, DE-PaLM2) and CorrW(AE, DE-PaLM2) show higher correlations with human judgments than DE-PaLM2, suggesting that assistant evaluators can enhance an LLM's performance beyond its standalone capabilities. However, Fusion-Eval surpasses these aggregation methods, making it better at leveraging assistant evaluators over mere score aggregation.

The performance of FE-PaLM2 is higher than that of FE-PaLM2-NoPlan, suggesting that prompting the LLM with a plan is beneficial. This improvement could be attributed to the plan aiding the LLM in utilizing assistant evaluators. This finding aligns with G-Eval~\citep{liu2023gpteval}, which suggests intrinsic evaluation steps generated by planning LLMs enhance performance, especially in complex evaluation tasks. However, the LLM-generated plan used in our experiments is likely not optimal. Finding an `optimal plan' is nearly impossible due to the exponential complexity involved in combining criteria and assistant evaluators. We recognize the potential for hallucinations in LLM-generated plans and note that a human-created plan could also be employed with Fusion-Eval.

\subsubsection{Fusion-Eval Execution Time}
The Fusion-Eval framework maintains a manageable execution time because the assistant evaluators have minimal inference times compared to LLMs. Running all assistant evaluators (NLI, BLEURT, and SumBLEURT) on a SummEval example takes about 0.125 seconds on average. The evaluators are pre-trained, eliminating the need for additional training. Obtaining a Fusion-Eval result using PaLM2, based on assistant evaluator scores, takes about 7 seconds for a SummEval example and 11.7 seconds for a TopicalChat example. 
%\ls{For DE-PaLM2, processing times are 2.6 seconds for a SummEval example and 5 seconds for a TopicalChat example.}

\subsubsection{Correlations between Fusion-Eval And Assistant Evaluators} % \nw{I'm having trouble understanding the purpose of this section. Maybe we can meet about it?}

% Tables~\ref{tab:fusioneval_assistant_evaluator_summeval_correlation} and \ref{tab:fusioneval_assistant_evaluator_topicalchat_correlation} illustrate the correlation of assistant evaluators with FE-PaLM2.
% Similarly, Tables~\ref{tab:fusioneval_gpt4_assistant_evaluator_summeval_correlation} and \ref{tab:fusioneval_gpt4_assistant_evaluator_topicalchat_correlation} detail the correlation of assistant evaluators with FE-GPT-4.
To understand Fusion-Eval's execution, we analyzed the correlation between its scores and those of the assistant evaluators, alongside the evaluators chosen by the LLM's plan. Tables~\ref{tab:fusioneval_assistant_evaluator_summeval_correlation} and \ref{tab:fusioneval_assistant_evaluator_topicalchat_correlation} detail the correlation for FE-PaLM2, while Tables~\ref{tab:fusioneval_gpt4_assistant_evaluator_summeval_correlation} and \ref{tab:fusioneval_gpt4_assistant_evaluator_topicalchat_correlation} do the same for FE-GPT-4. The planning LLM's evaluator selections are listed in Table~\ref{tab:llm_suggested_assistant_evaluator}.

Across evaluation dimensions, the LLM's chosen evaluators consistently exhibit higher correlations with both FE-PaLM2 and FE-GPT-4 compared to those not selected. For instance, in SummEval's coherence, SumBLEURT demonstrates a higher correlation than other evaluators. A similar trend is also observed in TopicalChat's naturalness and understandability. This suggests Fusion-Eval does rely on selected assistant evaluators more than non-selected ones. 

Moreover, the absence of a perfect correlation (``$1$'') between Fusion-Eval and any assistant evaluator suggests that Fusion-Eval uses assistant evaluators to supplement its judgment rather than relying entirely on them.

\section{Conclusion}
\label{sec:conclusion}
The paper presents Fusion-Eval, an innovative aggregator using Large Language Models (LLMs) for diverse evaluation tasks. It effectively integrates assistant evaluators according to specific criteria. Empirical results show Fusion-Eval achieves higher correlations with human judgments than baselines. LLMs are very powerful, so it's interesting that augmenting LLMs with scores from simpler methods can improve performance in this case.
\newpage

\bibliography{colm2024_conference}
\bibliographystyle{colm2024_conference}
\newpage
\appendix
\section{Limitation and Future Work}
\label{sec:limitation}

The length of our execution prompt templates for SummEval (Appendix~\ref{sec:appendix:prompt_summeval}) and TopicalChat (Appendix~\ref{sec:appendix:prompt_topicalchat}) is 662 and 990 words, respectively. The LLMs used in Fusion-Eval, including GPT-4 and PaLM2, can effectively process prompts of this length. However, the lengthy Fusion-Eval prompts may present challenges for LLMs with limited context windows. To address this, we propose investigating prompt decomposition in future work to enhance Fusion-Eval's compatibility with various LLMs.

\section{Appendix}
\label{sec:appendix}
% \subsection{Evaluation Benchmark}
% SummEval~\cite{fabbri2021summeval} contains 1600 data points (16 summarization systems $\times$ 100 examples) in total. 
% TopicalChat~\cite{mehri2020usr} has 360 data points (6 dialogue response generation systems $\times$ 60 examples) in total.

% Details of calculations follow. \nw{I think we should remove this since it's redundant with the part we already explained, and more confusing.}
% \begin{itemize}
%     \item AVG(AE): Average scores of assistant evaluators for each example.
%     \item AVG(AE, DE-PaLM2 or G-Eval-GPT-4): For each example and dimension, average scores of assistant evaluators and DE-PaLM2 or G-Eval-GPT-4. For example, AVG(AE, DE-PaLM2)[`Coh'] = AVG(BLEURT, NLI, SumBLEURT, DE-PaLM2[`Coh']).
%     \item LLMSel(AE): Average scores of LLM-selected assistant evaluators for each example.
%     \item LLMSel(AE, DE-PaLM2 or G-Eval-GPT-4): Similar to AVG, but only using LLM-selected assistant evaluators with DE-PaLM2 or G-Eval-GPT-4 for each dimension in each example.
%     \item CorrW(AE): Weighted average of assistant evaluators' scores based on their correlation to human judgment. For example, CorrW(AE)[`Coh'] = (0.443*BLEURT + 0.45*NLI + 0.7*SumBLEURT) / (0.443 + 0.45 + 0.7).
%     \item CorrW(AE, DE-PaLM2 or G-Eval-GPT-4): A weighted average including DE-PaLM2 or G-Eval-GPT-4. For example, CorrW(AE,DE-PaLM2)[`Coh'] = (0.443*BLEURT + 0.45*NLI + 0.7*SumBLEURT + 0.733*DE-PaLM2[`Coh']) / (0.443 + 0.45 + 0.7 + 0.733).
% \end{itemize}

\subsection{Fusion-Eval Evaluation Prompt Template for SummEval (One Prompt Only in This Subsection - Do Not Be Surprised by Its Length)}
\label{sec:appendix:prompt_summeval}
\emph{Sections before the input template are generated by the planning LLM, while those after it are human-created.} 

\begin{lstlisting}[basicstyle=\small]
Evaluate a provided summary using criteria: Coherence, Consistency, Relevance, and Fluency.


Assistant Evaluators like NLI, BLEURT, and SUM_BLEURT, which give scores between below 0 and 1 (closer to 1 being better), will assist in this evaluation.
**1. NLI (Natural Language Inference)**:
This assistant evaluator provides a probability score indicating how much the summary (hypothesis) is entailed by the original news article (premise).
**Usage**:
- **Consistency Evaluation**: A high entailment probability indicates that the summary is factually aligned with the source text. Conversely, a low score might indicate discrepancies or hallucinated facts.

**2. BLEURT**:
This metric models human judgments. It gives a score indicating how closely the summary aligns with what human evaluators might consider a good summary given the source text.
**Usage**:
- **Relevance and Consistency Evaluation**: A high BLEURT score would suggest that the summary effectively captures the essential points of the source. A low score might indicate missing key points.

**3. SUM_BLEURT (Summarization BLEURT)**:
Fine-tuned on a summarization dataset, this assistant evaluator offers a more targeted approach to measuring the quality of summaries in the context of human judgments.
**Usage**:
- **Relevance and Coherence Evaluation**: Like BLEURT, but given its specialization in summarization, SUM_BLEURT could offer more precise insights into the relevance and coherence of the summary in relation to the source text.


**Plan Using Assistant Evaluators**:
1. **Read the News Article and Summary**: Begin with a manual reading to form an initial impression.
2. **Use NLI & BLEURT for Consistency**: Check both scores. High scores from both assistant evaluators will reaffirm the consistency of the summary.
3. **Use BLEURT & SUM_BLEURT for Relevance**: Check scores from both assistant evaluators. High scores would suggest a good summary in terms of relevance.
4. **Use SUM_BLEURT for Coherence**: Check SUM_BLEURT score. High scores would suggest a good summary in terms of coherence.
5. **Manual Evaluation for Fluency**: The assistant evaluators don't directly address fluency. You'll evaluate grammar, punctuation, and sentence structure manually.
6. **Final Judgment**: The assistant evaluators' outputs will inform and validate your evaluations, but the ultimate judgment will be based on the provided criteria and steps, with the assistant evaluators serving as supplementary aids.


**Criteria & Steps**:
1. **Coherence (1-5)**:
   - Read the news article and the summary.
   - Compare the summary to the article for clarity and logical order.
   - Use SUM_BLEURT scores as supplementary insights for coherence.
   - Assign a coherence score based on organization and structure.

2. **Consistency (1-5)**:
   - Use NLI & BLEURT to get scores.
   - Read the article and summary.
   - Compare factual details.
   - Assign a consistency score based on factual alignment.

3. **Relevance (1-5)**:
   - Use BLEURT & SUM_BLEURT to get alignment scores with human-like judgments.
   - Read both the article and summary.
   - Identify main points and coverage in the summary.
   - Assign a relevance score based on content importance and absence of redundancies.

4. **Fluency (1-5)**:
   - Evaluate the summary manually for grammar, punctuation, and sentence structure.
   - Assign a fluency score based on readability.

**Evaluation Summary (1-5)**:
Consider the scores from each criterion and their importance.
   - Derive an average score, ensuring the final score ranges between 1-5.
   - Provide overall comments on the summary.
   - Highlight strengths and areas needing improvement.


**Input Template**:
Source:
[Provide the source text here]

Answer:
[Provide the summary text here]

NLI Score (Source as Premise and Answer as Hypothesis):
[Provide NLI entailment probability score]

BLEURT Score (Source as Premise and Answer as Hypothesis):
[Provide BLEURT score]

SUM_BLEURT Score (Source as Premise and Answer as Hypothesis):
[Provide SUM_BLEURT score]


**Output Template**:
Criterias' Scores and Explanations:

Coherence
Score: [Your evaluation] Explanation: [Your explanation on evaluation]

Consistency
Score: [Your evaluation] Explanation: [Your explanation on evaluation]

Relevance
Score: [Your evaluation] Explanation:[Your explanation on evaluation]

Fluency
Score: [Your evaluation] Explanation: [Your explanation on evaluation]

Evaluation Summary:
Overall Score: [Your evaluation]
Explanation: [Your explanation on evaluation]

**Input Example**:
Source:
[[source]]

Answer:
[[summary]]

NLI Score (Source as Premise and Answer as Hypothesis):
[[nli_score_source_answer]]

BLEURT Score (Source as Premise and Answer as Hypothesis):
[[bleurt_score_source_answer]]

SUM_BLEURT Score (Source as Premise and Answer as Hypothesis):
[[sum_bleurt_score_source_answer]]


Evaluation (please follow Output Template and provide the evaluation result):<<eval_result>>
\end{lstlisting}

\subsection{Fusion-Eval Evaluation Prompt Template for TopicalChat (One Prompt Only in This Subsection - Do Not Be Surprised by Its Length)}
\label{sec:appendix:prompt_topicalchat}
\emph{Sections before the input template are generated by the planning LLM, while those after it are human-created.}

\begin{lstlisting}[basicstyle=\small]
You will be given a conversation between two individuals, followed by a potential response for the next turn in the conversation, which includes an interesting fact. Your task is to rate the responses on six metrics: Coherence, Engagingness, Naturalness, Groundedness, Understandability, and Overall Quality.


Assistant Evaluators' Descriptions and Usage:
**1. LM_PROB (Language Model Probability):**
- **Functionality**: LM_PROB provides a probability score, ranging from 0 to 1, indicating the likelihood that a given response would be generated by a language model, given the preceding conversation and fact.
- **Score Range**: 0 (least likely) to 1 (most likely).
- **Usage**:
  - **Naturalness Evaluation**: A higher probability score suggests that the response is more likely to occur naturally in human conversation, indicating greater naturalness.
  - **Understandability Evaluation**: Similarly, a higher probability can also imply that the response is more understandable within the given context, as it is more aligned with expected language patterns.

**2. BLEURT:**
- **Functionality**: BLEURT evaluates the quality of text generation by comparing the generated text (response) to a reference (conversation and fact). Its score range is 0 to 1, where higher scores indicate better alignment and quality.
- **Score Range**: 0 (poor alignment) to 1 (excellent alignment).
- **Usage**:
  - **Groundedness Evaluation**: A high BLEURT score indicates that the response accurately and relevantly utilizes the given fact, showing strong groundedness in the context of the conversation.


Plan Using Tools for Conversation Response Evaluation:
1. **Read the Conversation, Fact, and Response**: Begin with a careful reading of the provided materials to form an initial qualitative impression of the response in the context of the conversation and fact.
2. **Use LM_PROB for Naturalness and Understandability Evaluation**:
   - Apply LM_PROB to determine the probability that the response would be generated by a language model in the given context.
   - High probability scores from LM_PROB will indicate greater naturalness and understandability, as the response aligns well with expected language patterns.
3. **Use BLEURT for Groundedness Evaluation**:
   - Employ BLEURT to assess how accurately and relevantly the response utilizes the given fact in the context of the conversation.
   - A high score from BLEURT suggests that the response is well-grounded in the provided fact, demonstrating accuracy and relevance.
4. **Final Judgment and Integration of Tool Outputs**:
   - Integrate the outputs from the tools with your initial qualitative assessment.
   - The tools' outputs will provide quantitative support and validation for your evaluations in each metric.
   - Make the final judgment based on a holistic view, considering both the tool outputs and the original evaluation criteria for each metric.
   - Remember that the ultimate judgment should align with the predefined criteria and evaluation steps, with the tools serving as important but supplementary aids in the decision-making process.


**Criteria & Steps**:
1. **Coherence (1-3, Any Floating Value)**:
   - Read the conversation, fact, and response to assess the logical flow and continuity.
   - Evaluate how well the response connects with and continues the conversation.
   - Assign a Coherence score, ranging from 1 to 3, based on the response's organization and logical integration into the conversation.

2. **Engagingness (1-3, Any Floating Value)**:
   - Review the conversation, fact, and response to determine the level of interest or intrigue.
   - Assess how the response contributes to the conversation's value and captivates interest.
   - Assign an Engagingness score, ranging from 1 to 3, based on the response's ability to captivate and add value to the conversation.

3. **Naturalness (1-3, Any Floating Value)**:
   - Read the conversation, fact, and response to gauge the natural fit of the response within the conversation's context.
   - Evaluate the tone, formality, and conversational flow to determine how naturally the response fits.
   - Use LM_PROB to supplement the evaluation, considering the likelihood of such a response in the given context.
   - Assign a Naturalness score, ranging from 1 to 3, focusing on how naturally the response fits into the conversation.

4. **Groundedness (0-1, Any Floating Value)**:
   - Examine the conversation, fact, and response to evaluate how well the response utilizes the given fact.
   - Assess the accuracy and relevance of the fact in the response.
   - Utilize BLEURT to provide supplementary insights into how accurately the response is grounded in the given fact.
   - Assign a Groundedness score, ranging from 0 to 1, based on the effective and accurate incorporation of the fact in the response.

5. **Understandability (0-1, Any Floating Value)**:
   - Review the conversation, fact, and response to assess the clarity and comprehension of the response.
   - Focus on how clearly and easily the response can be understood within the context of the preceding conversation.
   - Apply LM_PROB for additional data on the understandability of the response.
   - Assign an Understandability score, ranging from 0 to 1, based on the response's clarity and ease of comprehension in context.

6. **Overall Quality (1-5, Any Floating Value)**:
   - Review the scores and insights from the previous criteria, including data from assistant evaluators.
   - Consider how the aspects of Coherence, Engagingness, Naturalness, Groundedness, and Understandability collectively contribute to the overall impression of the response.
   - Assign an Overall Quality score, ranging from 1 to 5, based on a holistic assessment of the response's strengths and weaknesses.
   - Provide a summary explanation for the overall quality rating, highlighting key factors and insights that influenced the judgment. 


**Input Template**:
Conversation:
[Provide the conversation text here]

Fact:
[Provide the fact text here]

Response:
[Provide the response text here]

LM_PROB Score (Response in Context of Conversation and Fact):
[Provide LM_PROB probability score]

BLEURT Score (Response with Conversation and Fact as Reference):
[Provide BLEURT score]


**Output Template**:
Criteria Scores and Explanations:

Coherence
Score: [Your evaluation] Explanation: [Your explanation on evaluation]

Engagingness
Score: [Your evaluation] Explanation: [Your explanation on evaluation]

Naturalness
Score: [Your evaluation] Explanation: [Your explanation on evaluation]

Groundedness
Score: [Your evaluation] Explanation: [Your explanation on evaluation]

Understandability
Score: [Your evaluation] Explanation: [Your explanation on evaluation]

Evaluation Summary:
Overall Score: [Your evaluation] Explanation: [Your comprehensive explanation on the overall evaluation, integrating aspects from each criterion]


**Input Example**:
Conversation:
[[conversation]]

Fact:
[[fact]]

Response:
[[response]]

LM_PROB Score (Response in Context of Conversation and Fact):
[[lm_prob_score]]

BLEURT Score (Response with Conversation and Fact as Reference):
[[bleurt_score]]


Evaluation (please follow Output Template and provide the evaluation result):<<eval_result>>
\end{lstlisting}

% \begin{strip}
% \captionof{lstlisting}{GPT4 Generated Summeval Reference Based Form.}\label{lst:summeval_reference_based_prompt}

% \end{strip}

\end{document}